\documentclass{article} 
\usepackage[final]{colm2026_conference}

\usepackage{microtype}
\usepackage{hyperref}
\usepackage{url}
\usepackage{booktabs}

\usepackage{algorithm}
\usepackage{algorithmic}
\usepackage{amsmath}

\usepackage{lineno}

\usepackage{graphicx}
\usepackage{amsmath}
\usepackage{multirow}
\usepackage{wrapfig}

\usepackage{amssymb}
\usepackage{hyperref}
\usepackage{url}
\usepackage{booktabs}
\usepackage{subcaption}  
\usepackage{array}       
\usepackage{lipsum} 
\usepackage[table]{xcolor}
\usepackage{listings}

\lstdefinestyle{pseudocodefig}{
    basicstyle = \ttfamily\scriptsize,
    keywordstyle = \color{blue},
    commentstyle = \color{green!60!black},
    stringstyle = \color{red},
    numbers = left,
    numberstyle = \tiny\color{gray},
    numbersep = 4pt,
    firstnumber = 1,
    xleftmargin = 2.2em,
    framexleftmargin = 1.6em,
    frame = single,
    frameround = tttt,
    backgroundcolor = \color{gray!5},
    breaklines = true,
    breakatwhitespace = true,
    columns = flexible,
    keepspaces = true,
    language = Python,
    escapeinside = {(*@}{@*)},
}

\usepackage{bm}

\usepackage{multicol}
\definecolor{lightgray}{gray}{0.92}  

\newcommand{\name}{RL-VLA$^3$}

\definecolor{darkblue}{rgb}{0, 0, 0.5}
\hypersetup{colorlinks=true, citecolor=darkblue, linkcolor=darkblue, urlcolor=darkblue}

\title{RL-VLA$^3$: A Flexible and Asynchronous Reinforcement Learning Framework for VLA Training}

\author{Haoran Sun$^{1}$\thanks{Equal contributions, any order is acceptable by all authors.}, Yongjian Guo$^{2}$\footnotemark[1], Zhong Guan$^{3}$\footnotemark[1], \\
\textbf{Shuai Di$^{4}$, Xiaodong Bai$^{4}$, Jing Long$^{1,4}$, Tianyun Zhao$^{2}$, Mingxi Luo$^{4}$}, \\
\textbf{Xiaotie Deng$^{1}$, Xu Chu$^{1\dagger}$, Hongke Zhao$^{3\dagger}$, Likang Wu$^{3}$, Xi Xiao$^{2}$, Sheng Wen$^{2}$}, \\
\textbf{Yicheng Gong$^{4}$, Junwu Xiong$^{4}$}\thanks{Corresponding authors. (\texttt{chu\_xu@pku.edu.cn}, \texttt{hongke@tju.edu.cn}, 
\texttt{xiongjunwu.1@jd.com})}\\
$^1$Peking University, $^2$Tsinghua University, $^3$Tianjin University, $^4$JDT AI Infra \\$^5$Swinburne University of Technology\\
}

\begin{document}

\ifcolmsubmission
\linenumbers
\fi

\maketitle

\begin{abstract}
Reinforcement learning (RL) has emerged as a critical paradigm for post-training Vision-Language-Action (VLA) models, enabling embodied agents to adapt and improve through environmental interaction. 
However, existing RL frameworks for VLAs inherit synchronous design principles from traditional LLM training, treating entire rollouts as indivisible units and alternating strictly between data collection and policy optimization.
This fundamentally mismatches the unique characteristics of VLA training, as physical simulators introduce highly variable, resource-intensive latencies. 
To address this, we introduce RL-VLA$^3$, a fully asynchronous distributed RL framework that enables fine-grained asynchronous interaction between simulation, inference, and training components through dynamic batching schedulers and flexible environment sharding strategies. 
Extensive experiments across diverse simulation backends, VLA architectures, and RL algorithms demonstrate that RL-VLA$^3$ achieves throughput improvements of up to 85.2\% over synchronous baselines while maintaining identical sample efficiency, with scalability validated from 8 to 256 GPUs. 
To our knowledge, RL-VLA$^3$ is the first fully asynchronous RL training framework tailored specifically for the system-level challenges of VLA training.
\end{abstract}

\section{Introduction}\label{sec:intro}

Vision-Language Action models (VLAs)~\citep{ma2024survey,zhong2025survey,zhang2025pure} have emerged as a powerful paradigm for embodied AI, enabling agents to perceive and interact with their environments through a combination of visual and linguistic understanding. 
While foundation VLAs~\citep{zitkovich2023rt2,black2024pi_0,bjorck2025gr00t,gemini2025robotics,kim2024openvla,shukor2025smolvla} trained via large-scale supervised fine-tuning have demonstrated basic capabilities, reinforcement learning (RL) has shown promise in further enhancing their performance and adaptability~\citep{wagenmaker2025steering,guo2025improving,liu2025can,lu2025vla}. 
Currently, the predominant approach for training VLAs with RL relies on traditional RL frameworks that treat entire rollouts as indivisible units for policy optimization~\citep{li2025simplevla,zang2025rlinf,li2025vlarft}. 
Consequently, the underlying training infrastructure largely follows the same design principles as traditional RL frameworks for LLM agents.

However, \emph{VLA training exhibits distinct characteristics compared to classic LLM RL training}. 
Typical RL training can be divided into two phases: data generation (rollout) and policy optimization (training)~\citep{sheng2025hybridflow,hu2024openrlhf}. 
The rollout phase in LLM training is performed by an inference engine to generate large amounts of tokens, while VLA training involves extremely frequent interaction between the agent and the environment. 
Unlike environments such as web search or API calling, the environment in VLA training is a complex physics simulator, which is considerably more resource-intensive and time-consuming. 
For instance, in ManiSkill~\citep{mu2021maniskill}, a $\pi_{0.5}$ model~\citep{black2024pi_0} takes $9$ seconds to produce a batch of $640$ actions, while the simulator requires $22$ seconds to compute the next states and observations according to these actions. 
Moreover, different environment simulators exhibit varying patterns of resource consumption. 
Some simulators can efficiently leverage GPU resources for parallel processing~\citep{mu2021maniskill,nasiriany2024robocasa}, while others rely more heavily on CPU computations, resulting in varying levels of GPU utilization~\citep{liu2023libero,chen2025robotwin}. Still others involve a mix of CPU and GPU computations for rendering and physics simulation, leading to unstable GPU utilization~\citep{zhou2026thousand}.

\begin{figure}[t]
    \centering
    \includegraphics[width=0.95\linewidth,trim=0 0 0 0,clip]{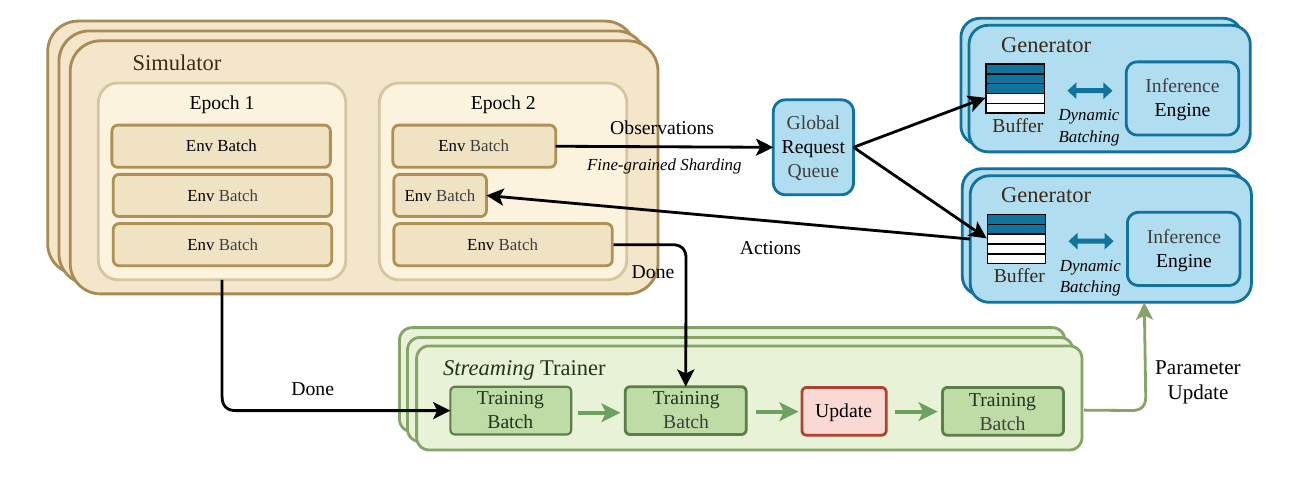}
    \caption{Overall architecture of \name. Generators, Simulators, and Trainers interact fully asynchronously; black arrows ($\to$) indicate data flow. We use \emph{fine-grained sharding} of environment batches and \emph{dynamic batching} for Generator inference to accommodate asynchronous interaction and rollouts, improving throughput.}    
    \label{fig:framework}
\end{figure}

Motivated by these observations, we propose \name\, a flexible \underline{R}einforcement \underline{L}earning framework specifically designed for training VL\underline{A} with \underline{A}synchronous rollout interaction and \underline{A}synchronous training. 
Following the RLinf~\citep{zang2025rlinf} framework, we explicitly define three resource groups: the Simulator, the Generator, and the Trainer. 
We introduce fine-grained decoupling, which provides a more flexible interface for users to configure.
The interaction logic between these three groups is illustrated in Figure~\ref{fig:framework} and will be detailed in Section~\ref{sec:methods-overall}.
Notably, \emph{our entire training process executes asynchronously, allowing all three resource groups to progress independently and simultaneously}.
The main contributions are as follows:
\begin{itemize}
    \item A fully flexible rollout interface that allows users to configure the grouping and number of parallel environments within the Simulator, and specify the interaction order between Simulators and Generators.
    \item A fully asynchronous training framework in which the three components, Generators, Simulators, and Trainers, interact entirely asynchronously, with user-configurable batching strategies for Generators.
    \item Extensive experimental evaluations demonstrating the training efficiency and performance improvements of \name\ across various tasks.
\end{itemize}
To our knowledge, \name\ is the first fully asynchronous distributed RL training framework specifically designed for VLAs.

The remainder of the paper is organized as follows.
In Section~\ref{sec:related}, we review related work on foundation VLA models and reinforcement learning frameworks for LLMs. 
In Section~\ref{sec:methods}, we present the design principles of \name, including the overall framework, asynchronous environment interaction, and asynchronous policy optimization. 
In Section~\ref{sec:experiments}, we report experimental results on training throughput and performance, along with ablation studies and resource-utilization analysis. 
Finally, in Section~\ref{sec:conclusion}, we conclude the paper and discuss future work.

\section{Related Work}\label{sec:related}

\subsection{Reinforcement Learning for Foundation VLAs}
Foundation VLAs models, such as RT-2~\citep{zitkovich2023rt2}, OpenVLA~\citep{kim2024openvla}, $\pi_0$~\citep{black2024pi_0}, and GR00T~\citep{bjorck2025gr00t}, have demonstrated strong basic robotic manipulation capabilities through large-scale supervised fine-tuning (SFT) on human-teleoperated data. However, because SFT policies often struggle to recover from out-of-distribution errors or generalize to novel states, reinforcement learning has emerged as a crucial post-training paradigm to enable continuous improvement and environmental adaptation~\citep{wagenmaker2025steering,guo2025improving,tan2025interactive,kim2025fine,xiao2025selfimprovingvisionlanguageactionmodelsdata}. 

Recent works have demonstrated the efficacy of RL in enhancing VLA success rates~\citep{liu2025can,lu2025vla,li2025vlarft,chen2025conrft}. To facilitate this, initial distributed RL frameworks like SimpleVLA~\citep{li2025simplevla} and RLinf~\citep{zang2025rlinf} have been proposed to adapt RL algorithms for embodied AI. 
Current frameworks typically require all GPUs to load all resource groups concurrently (a colocated strategy), which introduces substantial efficiency losses due to frequent context switching between resources, such as alternating back and forth between the physical simulator and the model inference engine. Although RLinf~\citep{zang2025rlinf} has proposed separated and hybrid resource allocation strategies to mitigate this overhead, they still rely on synchronous execution during the rollout phase and maintain a synchronous training-inference pipeline. These synchronization barriers severely bottleneck overall throughput. 
Motivated by these limitations, \name\ introduces what is, to the best of our knowledge, the first fully asynchronous RL training framework specifically designed for VLAs.

\subsection{Distributed RL Frameworks for LLMs vs. Embodied AI}
The rapid evolution of RL from Human Feedback (RLHF) has driven the development of highly efficient distributed RL frameworks for LLMs. Systems such as VeRL~\citep{sheng2025hybridflow} and OpenRLHF~\citep{hu2024openrlhf} introduce hierarchical control, zero-redundancy model resharding, and highly optimized vLLM inference to balance flexibility and efficiency. To overcome global synchronization bottlenecks in large-scale deployment, recent systems like AReaL~\citep{fu2025areal}, Laminar~\citep{sheng2025laminar}, and ROLLART~\citep{gao2025rollart} advocate for fully decoupled, fine-grained asynchronous execution at the trajectory level. 

While \name\ is inspired by these decoupled LLM architectures, VLA training presents fundamentally different system-level challenges. 
In LLM RL pipelines, the ``environment'' is typically a neural reward model that runs entirely on GPUs with stable, predictable latencies. Conversely, the ``environment'' in embodied AI is a complex physics simulator (e.g., ManiSkill~\citep{mu2021maniskill}, LIBERO~\citep{liu2023libero}). These simulators rely heavily on varied CPU and GPU computations, require substantial memory footprints, and exhibit highly unpredictable latencies due to dynamic collision calculations and rendering tasks. Simply migrating LLM RL frameworks to VLA tasks fails to address these simulator-specific bottlenecks. 
To address this, \name\ introduces a fully flexible, asynchronous simulator interface alongside dynamic batching strategies, specifically tailored to absorb the latency fluctuations of physics engines and maximize throughput for embodied AI tasks.

\section{Design Principles of \name}\label{sec:methods}

In this section, we elaborate on the system design of \name. We first provide an overview of the framework, defining the distinct resource groups, their interactions, data flow, and user-configurable features. Next, we delve into the specific mechanisms driving the two critical interactions within our pipeline: the asynchronous rollout between parallel Simulators and Generators, and the asynchronous training between the rollout workers (Simulators and Generators) and the Trainer.

\subsection{Overall Framework}\label{sec:methods-overall}

The standard reinforcement learning pipeline consists of two phases: data generation (rollout) and policy optimization (training). In synchronous RL, these phases execute sequentially, an iterative process expressed as $(\text{Rollout} \rightarrow \text{Training})^{N}$, where $N$ denotes the number of training steps. Because VLAs rely on physical simulation, the rollout phase can be further subdivided into Simulator-side environment stepping (which yields observations) and Generator-side action inference, expressed as $\text{Rollout} = (\text{Simulator} \rightarrow \text{Generator})^{N_e}$, where $N_e$ is the number of environment interactions. Motivated by these naturally decoupled phases, we build upon \citet{zang2025rlinf} by explicitly separating the architecture into three distinct resource groups: the Simulator, the Generator, and the Trainer. As illustrated in Figure~\ref{fig:framework}, each Generator loads the VLA model and operates an independent inference engine; each Trainer is dedicated to policy gradient computations and parameter updates; and each Simulator hosts several environment batches (Env). Within each batch, vectorized environments execute concurrently to produce parallel observations.

\paragraph{The Execution Flow.} 
To manage the high-throughput flow of these observations, we introduce a highly configurable, fully asynchronous communication mechanism. Once generated, observations are processed through an optional sharding step and posted to a global request queue. Users can implement arbitrary \emph{sharding strategies} to shard the environment batch into multiple slots and explicitly route them to specific Generators. The default request queue operates as a priority queue sorted by slot arrival time and also supports custom sorting functions. A \emph{dynamic batching scheduler} manages the trigger conditions for Generator inference, allowing users to bound execution by maximum batch size and maximum wait latency. When an environment batch completes a full episode, the entire trajectory is asynchronously transmitted to the Trainer for continuous policy optimization. Crucially, \emph{our entire training process executes asynchronously, allowing all three resource groups to progress independently and simultaneously.} 

For further illustration, we summarize the execution flow in pseudo code, which is presented in the following Figure~\ref{fig:async_pseudocode}. 
The main $\mathrm{pipeline}$ (Left) initializes rollouts (Line 4) and then continuously collects completed trajectories for optimization (Line 6). 
It also manages model version synchronization (Line 11) when the number of update epochs meets the global condition. 
The rollout phase is mainly executed by the $\mathrm{collect\_rollout}$ function (Right). 
It first starts the Simulator processing (Line 3-9), in which the simulators are computing the observation data. 
And it also controls the triggering of Generator inference (Line 11-12) via the dynamic batching scheduler. 
These asynchronous interactions reduce the blocking time and significantly improve the overall training throughput.

\begin{figure}[htbp]
    \centering
    \begin{minipage}[t]{0.44\textwidth}
        \centering
        \begin{lstlisting}[style=pseudocodefig]
async def pipeline():
    rollout_version = 0
    local_updates = 0
    spawn(collect_rollout(rollout_version))
    while not finished():
        batch = await queue.get()
        train(batch)
        local_updates += 1
        if local_updates == sync_interval:
            rollout_version += 1
            emit(sync_to_rollout(
                rollout_version))
            local_updates = 0
        \end{lstlisting}
    \end{minipage}
    \hfill
    \begin{minipage}[t]{0.55\textwidth}
        \centering
        \begin{lstlisting}[style=pseudocodefig]
async def collect_rollout(version):
    for env in environments:
        spawn(
            while not done(env):
                submit(env.state, version)
                action = await next_action(env)
                transition, env.state = env.step(action)
                await queue.put((transition, version))
                )
    while True:
        reqs = await collect_requests(
            max_batch_size, timeout_ms)
        return_actions(reqs, model.forward(reqs))
        \end{lstlisting}
    \end{minipage}
    \caption{Pseudocode of two asynchronous components:
             \texttt{pipeline} (left) manages versioned training and synchronization,
             and \texttt{collect\_rollout} (right) handles environment stepping, request submission, and batched inference.}
    \label{fig:async_pseudocode}
\end{figure}

In the following two subsections, we detail the specific mechanisms enabling these core interactions: asynchronous environment stepping between the Simulator and Generator, and asynchronous policy optimization between the rollout workers and the Trainer.


\begin{figure}
    \centering
    \includegraphics[width=1\linewidth]{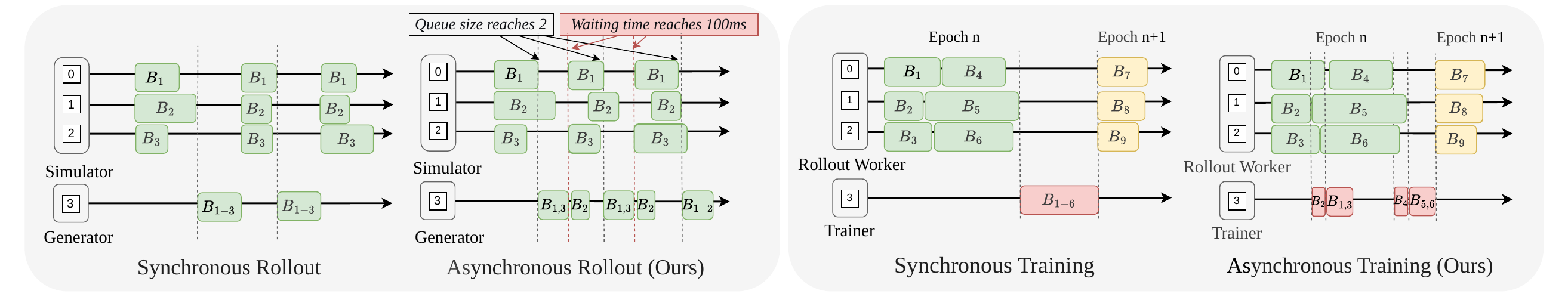}
    \caption{A demonstration of our proposed asynchronous Rollout (Left) and Training (Right). We introduce a dynamic batching scheduler to aggregate requests into a batch and trigger Generator inference.}
    \label{fig:async_rollout_and_train}
\end{figure}

\begin{figure}[t]
    \centering
    \includegraphics[width=0.99\linewidth,trim=21 0 21 0,clip]{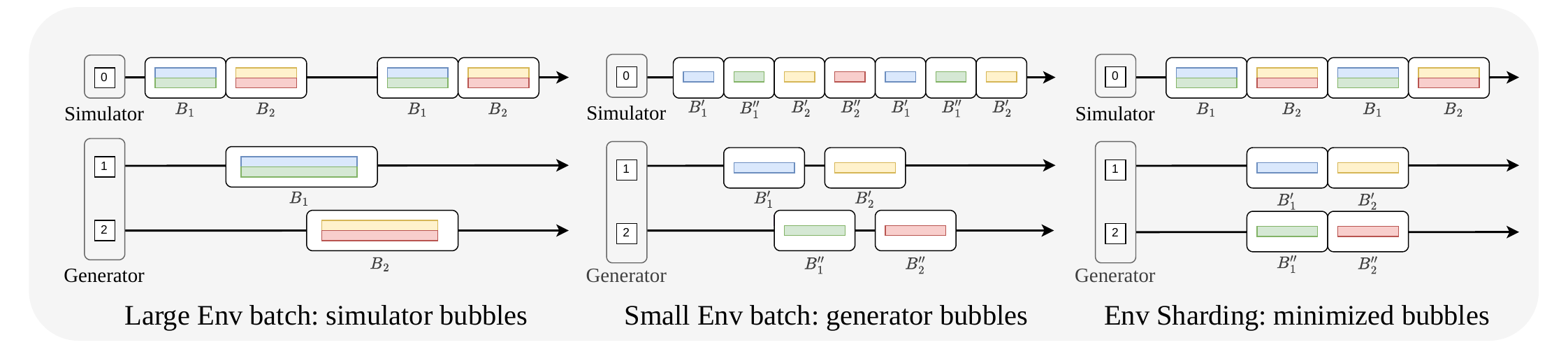}
    \caption{Fine-grained environment sharding for parallel interaction. Our framework shards parallel environment batches into multiple slots and distributes requests across different Generators. This maintains high throughput from large environment batches while reducing the Simulators' waiting time.}
    \label{fig:sharding_env}
\end{figure}

\subsection{Asynchronous Rollout between Parallel Simulators and Generators}\label{sec:methods-async-env}

During the RL training of VLA models, rollout throughput is heavily bottlenecked by rigid synchronization dependencies. As illustrated in Figure~\ref{fig:async_rollout_and_train} (Left), classic synchronous rollout frameworks wait for all Simulators to finish environment stepping before collecting observations to form a unified inference batch for the Generator. This creates severe synchronization overhead and exacerbates the long-tail latency caused by slower, computationally heavy environment batches. 

To resolve this, we propose a fine-grained asynchronous interaction mechanism that strictly decouples Simulator stepping from Generator inference. Instead of relying on global synchronization barriers, observation requests are posted to a request queue immediately after an individual environment completes its step. Generators are then triggered to perform inference independently, without waiting for the entire cohort of Simulators to finish. To maximize throughput and adapt to the varying computational characteristics of different environments, we introduce two flexible strategies: a \emph{dynamic batching scheduler} and \emph{fine-grained environment sharding}.

\paragraph{Dynamic Batching Scheduler.}
Because most environments do not natively support generating massive observation batches instantaneously, triggering Generator inference the moment a queue becomes non-empty leads to severe hardware underutilization. To counter this, we introduce a dynamic batching scheduler to intelligently aggregate requests into an optimal batch before inference. As shown in Figure~\ref{fig:async_rollout_and_train} (Right), the scheduler operates on two orthogonal constraints: a maximum batch size and a maximum wait latency. Generator inference is triggered only when the queue size reaches the batch size threshold, or when the oldest pending request exceeds the latency limit. By tuning these hyperparameters, we can effectively eliminate Generator pipeline bubbles and improve overall throughput via optimized inference batch sizes.

\paragraph{Fine-grained Environment Sharding.}
For environments capable of a high degree of parallelism, we shard massive environment batches into multiple smaller slots, distributing the resulting requests across different Generators. This strategy preserves the high throughput inherent to large environment batches while drastically reducing the waiting time for any single Simulator. As demonstrated in Figure~\ref{fig:sharding_env}, mapping a massive environment batch to a single Generator forces the Simulator to stall while waiting for inference, creating Simulator-side bubbles. Conversely, using undersized batches fails to saturate the Simulators' throughput, creating Generator-side bubbles. By sharding batches and routing slots across multiple Generators, our framework strikes an optimal balance, effectively minimizing idle time across both resources.

\subsection{Asynchronous Training}\label{sec:methods-async-optimization}

Beyond the asynchronous rollout, we also decouple the global interaction between the rollout workers and the Trainer to eliminate hardware idle time during policy updates. In standard synchronous RL pipelines (Figure~\ref{fig:async_rollout_and_train}, Right), data collection and policy optimization alternate strictly. The Trainer must wait for all rollout workers to complete their assigned trajectories and aggregate a full dataset before initiating the epoch's policy update. This rigid barrier inevitably leaves the Trainer idle during the rollout phase and forces the rollout workers to stall during the training phase.

To eliminate these pipeline bubbles, we implement a continuous, fully asynchronous training mechanism. In \name, once a Simulator batch completes an episode, the resulting trajectory is immediately pushed to the Trainer. The Trainer continuously processes this incoming data, ensuring that both the rollout workers and the optimization engine maintain high utilization. This overlapping execution significantly reduces the overall wall-clock training time.

\section{Experimental Results}\label{sec:experiments}

In this section, we evaluate \name\ against baseline training strategies across a diverse set of configurations, encompassing various backbone models, simulation environments, training algorithms, and GPU resource scales.\footnote{The code is available at \url{https://github.com/Haoran0301/RL-VLA3}} Our experiments demonstrate that our proposed flexible, asynchronous execution framework achieves substantially higher throughput while maintaining strict training stability.

\subsection{Experiment Setup}\label{sec:experiments-setup}

\paragraph{Models, Simulators, and Algorithms.}
We evaluate \name\ using several state-of-the-art pretrained VLA models. This includes diffusion-based architectures such as GR00T N1.5~\citep{bjorck2025gr00t} and the $\pi$ series ($\pi_0$ and $\pi_{0.5}$)~\citep{black2024pi_0}, as well as OpenVLA-OFT~\citep{kim2025fine}, an autoregressive model. Following \citet{zang2025rlinf}, we adapt OpenVLA-OFT to predict action chunks rather than single-step actions to improve interaction efficiency.

For simulation environments, we select LIBERO~\citep{liu2023libero}, ManiSkill~\citep{mu2021maniskill}, Meta-World~\citep{yu2020meta}, and RoboCasa~\citep{nasiriany2024robocasa}. This selection provides a comprehensive testbed of simulation backends: LIBERO and Meta-World rely on the CPU-bound MuJoCo physics engine; ManiSkill utilizes highly parallelized, GPU-accelerated ray-traced rendering; and RoboCasa features computationally heavy, photorealistic environments based on Robosuite. 
For policy optimization, we test our framework using two representative reinforcement learning algorithms: Proximal Policy Optimization (PPO)~\citep{schulman2017proximal} and Group Relative Policy Optimization (GRPO)~\citep{shao2024deepseekmath}.

\paragraph{Baselines and Evaluation Metrics.}
We build our codebase upon the foundation of RLinf~\citep{zang2025rlinf} and adopt their synchronous training pipeline as our primary baseline. We compare performance across two GPU placement strategies: \textbf{Colocated}, where a single GPU timeshares the Simulator, Generator, and Trainer; and \textbf{Hybrid}, where each GPU hosts a Trainer alongside either a Simulator or a Generator.

To assess system efficiency, our primary metric is \textbf{throughput}, defined as the total number of environment state transitions processed per unit time. Assuming a constant action chunk size, this is mathematically equivalent to the total number of action-inference steps executed by the Generator per unit time. To ensure a rigorous and fair architectural comparison, we carefully calibrate hyperparameters to maintain comparable overall GPU utilization between the baseline and \name, thereby isolating the performance gains attributed strictly to our asynchronous design. Detailed configurations are provided in Section~\ref{app:experiments}. To verify algorithmic correctness, we also evaluate \textbf{training performance} by plotting success rates over a finite step count.

\subsection{Benchmark Results}\label{sec:experiments-throughput}

In this subsection, we present the main benchmark results demonstrating the efficiency and scalability of \name.

\begin{figure}[t]
    \centering
    \includegraphics[width=\linewidth]{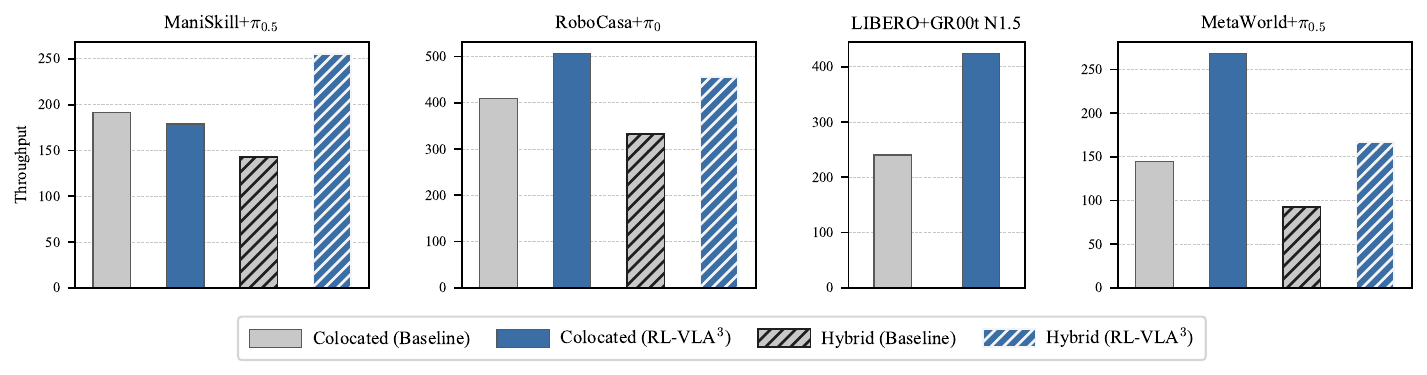}
    \caption{Single-node (8-GPU) throughput of \name\ versus the synchronous baseline across backbone models, environments, and the Colocated/Hybrid placements reported in our setup. \name\ achieves the highest throughput in every reported setting.}
    \label{fig:throughput_comparison}
\end{figure}

\begin{figure}[t]
    \centering
    \includegraphics[width=0.32\linewidth]{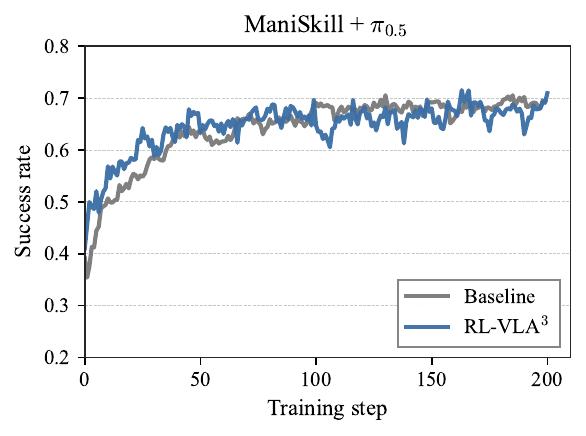}\hfill
    \includegraphics[width=0.32\linewidth]{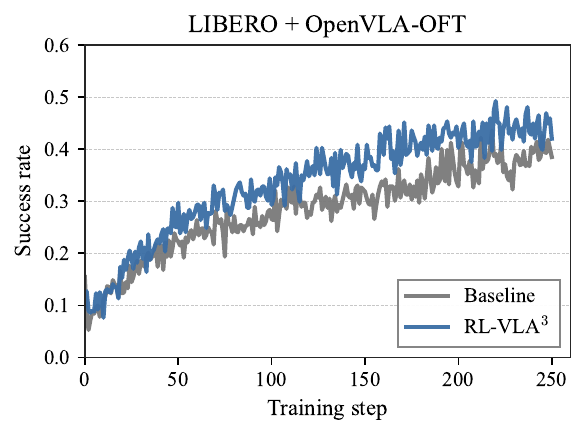}
    \includegraphics[width=0.32\linewidth]{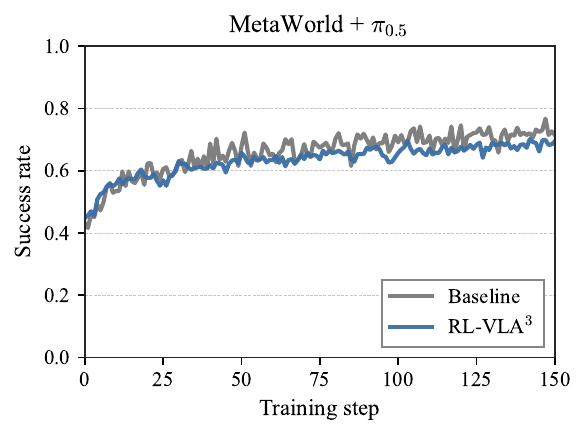}
    \caption{Success-rate curves for ManiSkill+$\pi_{0.5}$, LIBERO+OpenVLA-OFT, and Meta-World+$\pi_{0.5}$ under the synchronous baseline versus \name. The trajectories align in environment steps, confirming comparable sample efficiency while \name\'s higher throughput substantially reduces wall-clock time to comparable success levels.}
    \label{fig:success_rate}
\end{figure}

\paragraph{Throughput Comparison.}
Figure~\ref{fig:throughput_comparison} illustrates the throughput of \name\ compared to the synchronous baseline on a single 8-GPU node. Because LIBERO environments do not require GPU acceleration for physics computation, we omit Simulator-dedicated GPUs for LIBERO and report only Colocated placement. 

As shown, \name\ achieves the highest throughput across all environments. Under the Hybrid placement strategy, \name\ outperforms the baseline by approximately 78.6\% on ManiSkill, 36.7\% on RoboCasa, and 80.9\% on Meta-World. These substantial gains confirm that our fully asynchronous design successfully masks the blocking time inherent to the synchronous rollout and training phases. 

Under the Colocated placement strategy, \name\ still yields significant improvements: 23.5\% on RoboCasa, 76.3\% on LIBERO, and 85.2\% on Meta-World. The synchronous baseline in Colocated mode is heavily bottlenecked by fluctuating resource demands, resulting in inefficient GPU/CPU utilization. \name\ resolves this by enabling the Simulator, Generator, and Trainer to execute asynchronously, yielding a continuously smoothed, high-utilization workload. 

ManiSkill presents an interesting edge case. Because its simulation is natively highly parallelizable, the baseline strategy can already achieve massive batch sizes (up to 256) for both environment stepping and inference. In a strictly Colocated setup, forcing asynchronous execution can paradoxically introduce overhead due to hyper-frequent context switching between the GPU-heavy simulation and the model inference engine. However, \name\ easily overcomes this in Hybrid mode; by physically isolating the Simulator and Generator on separate GPUs, we preserve the massive environment batch sizes while eliminating the context-switching penalty, resulting in superior end-to-end throughput.

\paragraph{Scaling Behavior.}
We further investigate the distributed scaling properties of \name\ compared to the baseline. 
Figure~\ref{fig:scaling_law} plots the throughput for the LIBERO+GR00T N1.5 configuration scaling from 8 up to 256 GPUs on various placement modes. 
Here, we include a disaggregated mode which places the rollout workers (Generator and Simulator) and Trainer on separate GPUs.
\name\ exhibits near-linear scaling efficiency from 8 to 24 GPUs. While the scaling efficiency moderates up to 128 GPUs and experiences sublinear degradation toward 256 GPUs, \name\ maintains a consistent throughput advantage over the baseline at every scale. The efficiency drop at extreme scales is a known artifact of distributed RL, primarily driven by the heavy communication overhead required for weight broadcasting and gradient synchronization. Mitigating this communication bottleneck at the 256+ GPU scale remains a promising direction for future work.

\paragraph{Training Performance.}
To confirm that asynchronous execution does not degrade the mathematical integrity of the policy updates, we validate the training stability of \name. 
As shown in Figure~\ref{fig:success_rate}, \name\ achieves success rate curves directly comparable to the synchronous baseline with respect to training steps. 
Consequently, because our framework generates these steps at a significantly higher throughput, \name\ drastically accelerates the policy's wall-clock convergence time.

\subsection{Ablation Studies}\label{sec:experiments-ablation}

\begin{figure}[!t]
    \centering
    \begin{minipage}[t]{0.48\textwidth}
        \centering
        \includegraphics[width=\linewidth]{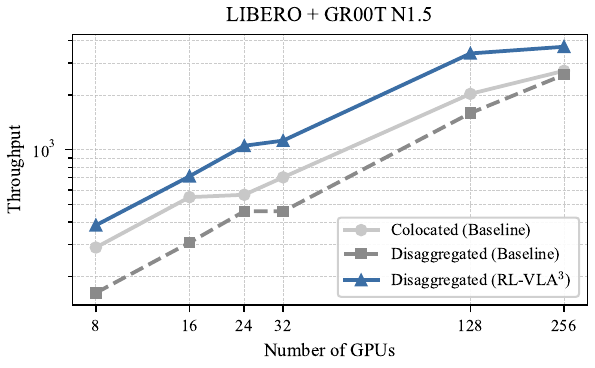}
        \captionof{figure}{The scaling behavior regarding the throughput versus GPU count for LIBERO+GR00T N1.5 under multiple placement modes. 
        }
        \label{fig:scaling_law}
    \end{minipage}
    \hfill
    \begin{minipage}[t]{0.48\textwidth}
        \centering
        \includegraphics[width=\linewidth]{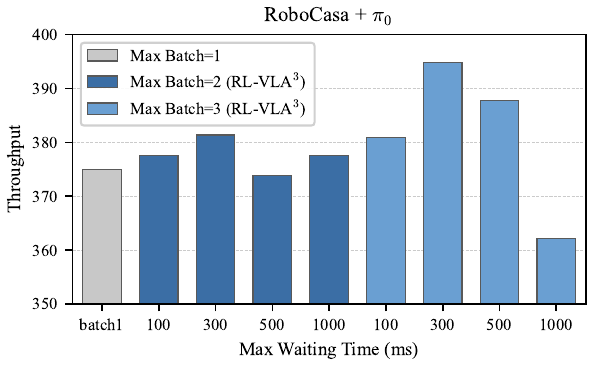}
        \captionof{figure}{RoboCasa+$\pi_0$ throughput under Hybrid placement when varying the dynamic batching scheduler's maximum batch number and latency threshold. 
        }
        \label{fig:robocasa_dynamic_batching}
    \end{minipage}
\end{figure}

\paragraph{Ablation on Dynamic Batching Strategy.}
To validate the dynamic batching scheduler introduced in Section~\ref{sec:methods-async-env}, we analyze its impact on the throughput using the RoboCasa + $\pi_0$ configuration in Hybrid placement. 
We allocate 6 GPUs for the Simulators and 2 GPUs for the Generators, assigning two parallel environment batches to each Simulator GPU. Because the sizes of these 12 environment batches are identical, we define the "maximum batch number" as the proxy for maximum batch size. 

The results, presented in Figure~\ref{fig:robocasa_dynamic_batching}, illustrate a classic systems trade-off between batching efficiency and queueing latency. Taking a maximum batch number of 3 as an example, throughput initially climbs as the maximum tolerable latency increases, before eventually declining. When the latency threshold is too restrictive, the system is forced into suboptimal, small-batch inference. Conversely, an excessively high latency threshold introduces severe queueing delays that stall the pipeline and outweigh the computational benefits of larger batches. This ablation confirms that properly tuning the dynamic batching hyperparameters is critical for locating the optimal hardware operating point.

\paragraph{Ablation on Different Levels of Asynchrony.}
Finally, we isolate the independent contributions of rollout-side and training-side asynchrony, which are denoted by Rollout Async and Train Async, respectively. 
Table~\ref{tab:ablation_level_on_async} reports throughput for four settings: (a) LIBERO+$\pi_{0.5}$ in hybrid placement; (b) OpenVLA-OFT+$\pi_{0.5}$ in hybrid placement; (c) RoboCasa+$\pi_{0}$ in colocated placement; and (d) MetaWorld+$\pi_{0.5}$ in hybrid placement, and various GPU scales from 8 up to 32.

\begin{table}[t]
\centering
\renewcommand{\arraystretch}{1.3}
\caption{Ablation study on different levels of asynchrony. We present the throughput for $4$ different configurations: (a) LIBERO+$\pi_{0.5}$ on Hybrid mode; (b) OpenVLA-OFT+$\pi_{0.5}$ on Hybrid mode; (c) RoboCasa+$\pi_{0}$ on Colocated mode; (d) MetaWorld+$\pi_{0.5}$ on Hybrid mode.}
\label{tab:ablation_level_on_async}
\resizebox{\textwidth}{!}{%
\begin{tabular}{lcccccccc}
\toprule
\multirow{2}{*}{Configuration} & \multicolumn{3}{c}{(a)} & \multicolumn{3}{c}{(b)} & (c) & (d) \\
\cmidrule(lr){2-4} \cmidrule(lr){5-7} \cmidrule(lr){8-8} \cmidrule(lr){9-9}
 & 8 GPUs & 16 GPUs & 32 GPUs & 8 GPUs & 16 GPUs & 32 GPUs & 8 GPUs & 8 GPUs \\
\midrule
Baseline & 162.75 & 307.84 & 457.23 & 144.35 & 249.66 & 472.33 & 379.25 & 92.29 \\
+ Rollout Async & 229.68 & 441.49 & 737.46 & 242.29 & 397.19 & 748.98 & 437.60 & 113.62 \\
+ Rollout \& Train Async & 383.40 & 713.38 & 1120.91 & 376.31 & 569.88 & 989.22 & 505.68 & 166.94 \\
\bottomrule
\end{tabular}%
}
\end{table}

In every configuration, isolating either asynchronous component yields measurable throughput improvements over the baseline. 
Notably, Train Async consistently provides the more pronounced performance lift. This suggests that successfully overlapping the computationally expensive policy gradient updates with continuous data generation is the dominant factor in maximizing end-to-end throughput. 
Rollout Async also shows clear improvements; we attribute these primarily to our dynamic batching strategy for environment interaction (Section~\ref{sec:methods-async-env}), and the consistent lift over the baseline further validates the effectiveness of our asynchronous design.

\subsection{Resource Utilization Analysis}\label{sec:experiments-resource}

\begin{figure}[t]
    \centering
    \includegraphics[width=0.95\linewidth]{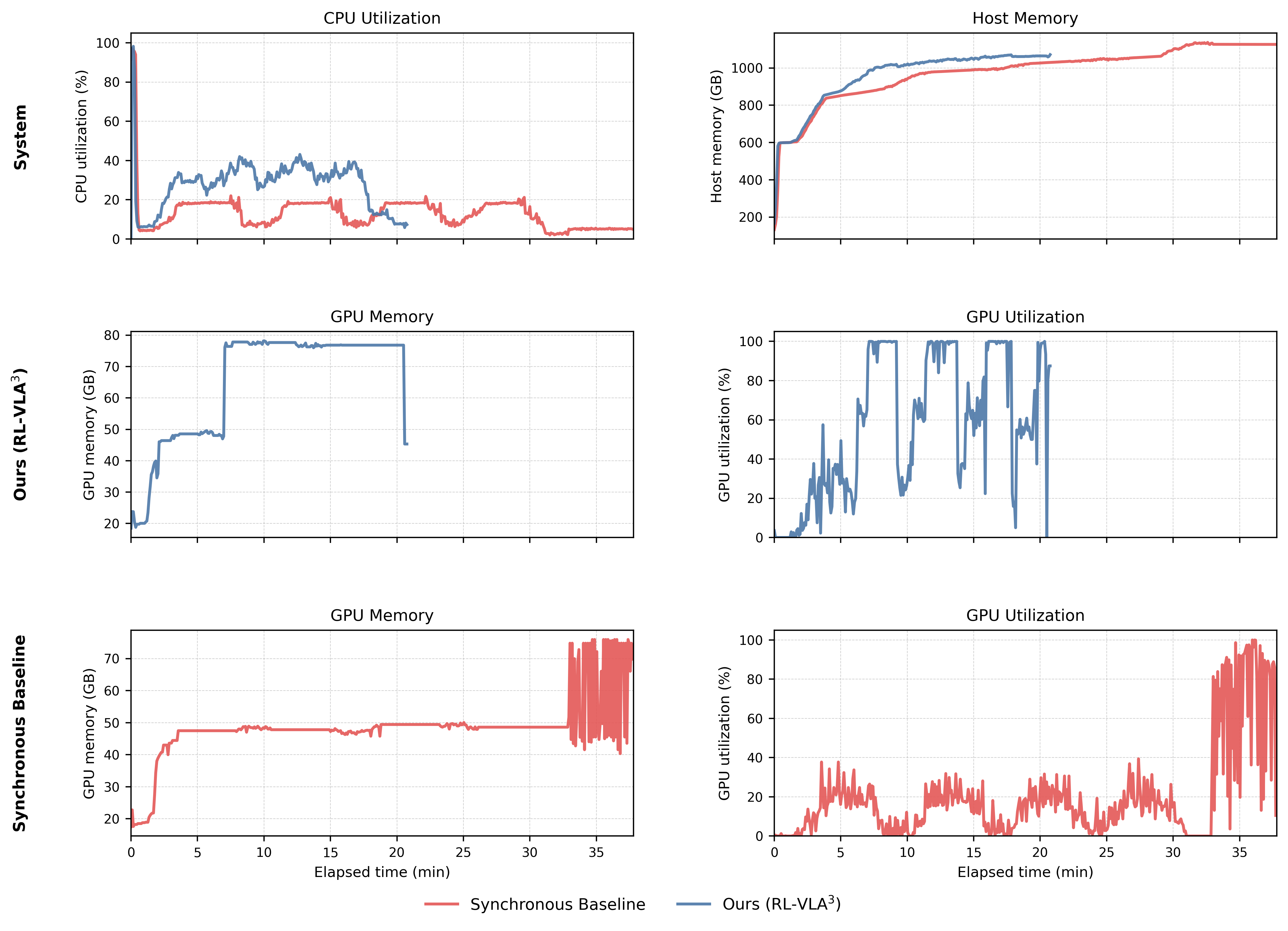}
    \caption{Comparison of resource utilization during the first training step between Colocated (Baseline) and Colocated (Ours) on LIBERO+GR00T N1.5. \name\ finishes the step substantially earlier while sustaining higher GPU utilization, as training begins once the first rollout completes rather than after the full rollout cohort.}
    \label{fig:resource-libero}
\end{figure}

Beyond aggregate throughput, we examine fine-grained CPU/GPU utilization and memory usage during the first training step to make the pipeline-level gains more concrete. 
Figure~\ref{fig:resource-libero} compares the synchronous baseline and \name\ on LIBERO+GR00T N1.5 under colocated placement, the setting that achieves the highest throughput for each method in Table~\ref{tab:app_libero}.

The utilization curves reveal two characteristic differences. 
First, \name\ starts policy optimization immediately after the first rollout finishes, so GPU memory and utilization rise early and remain elevated; the baseline keeps the GPU comparatively idle until all rollouts are collected and then incurs a late training burst. 
Second, asynchronous training streams trajectories in batches, which avoids the one-shot host-to-device transfer of a large VLA dataset (including image observations) that drives prolonged GPU waiting and host-memory pressure in the synchronous pipeline. 
Together, these effects explain why the same colocated hardware finishes the first step much faster under \name. 
Full utilization traces for ManiSkill and Meta-World, along with additional discussion, are provided in Appendix~\ref{app:resource-utilization}.

\section{Conclusion and Future Work}\label{sec:conclusion}

In this paper, we introduced \name, a fully asynchronous distributed reinforcement learning framework designed specifically for training Vision-Language-Action models. By enabling fine-grained asynchronous interaction between simulation, inference, and training components through dynamic batching schedulers and flexible environment sharding strategies, we eliminate the rigid synchronization barriers that plague traditional synchronous RL frameworks. Extensive experiments demonstrate that \name\ achieves substantial throughput improvements of up to 85.2\% over synchronous baselines while maintaining identical sample efficiency, with scalability validated from 8 to 256 GPUs. 

Several promising directions remain for future exploration. First, optimizing communication overhead between resource groups could further improve scaling efficiency, particularly at extreme scales where current bottlenecks limit sharding strategies. Second, developing autonomous agents to dynamically adjust placement strategies, batching scheduler hyperparameters, and environment sharding policies represents an exciting avenue for adaptive system optimization. Third, designing more efficient RL algorithms specifically tailored to the unique characteristics of VLA training could yield both computational and sample efficiency gains.


\section*{Acknowledgements}
This work was supported by New Generation Artificial Intelligence-National Science and Technology Major Project (2025ZD0122904), Natural Science Foundation of China (Grant No. 62572010), the Beijing-Tianjin-Hebei Natural Science Foundation Cooperation Project (25JJJJC0045), Natural Science Foundation of Tianjin (No. 24JCQNJC01560), Natural Science Foundation of Guangdong Province (2025A1515011946).

\section*{Ethics Statement}
This paper presents a novel framework for reinforcement learning in virtual learning environments. We have taken care to ensure that our research adheres to ethical guidelines, including considerations for data privacy, fairness, and potential societal impacts.

\bibliography{colm2026_conference}
\bibliographystyle{colm2026_conference}

\appendix

\appendix



\section{Further Experimental Results}\label{app:experiments}

In this section, we provide additional experimental results on more placement strategies and hyperparameters.

\begin{table*}[!h]
\centering
\caption{Detailed hyperparameters and throughput for different experimental setups on ManiSkill+$\pi_{0.5}$ with GRPO.}
\label{tab:app_maniskill}
\scriptsize

\renewcommand{\arraystretch}{1.25}

\begin{tabular*}{\textwidth}{@{\extracolsep{\fill}} l cccccccc @{}}
\toprule
\textbf{Method} & Baseline & Baseline & \name\ & Baseline & \name\ & Baseline &  \name\ & \name\ \\
\midrule

\rowcolor{lightgray}\multicolumn{9}{l}{\textit{\textbf{Placement Settings}}} \\
\quad Simulator & 0-7 & 0-7 & 0-7 & 0-7 & 0-7 & 0-5 & 0-5 & 0-5 \\
\quad Generator & 0-7 & 0-7 & 0-7 & 0-7 & 0-7 & 6-7 & 6-7 & 6-7 \\
\quad Trainer & 0-7 & 0-7 & 0-7 & 0-7 & 0-7 & 0-7 & 0-7 & 0-7 \\
\midrule

\rowcolor{lightgray}\multicolumn{9}{l}{\textit{\textbf{Environment Settings}}} \\
\quad Total Env & 2048 & 2048 & 2048 & 2048 & 2048 & 3264 & 3264 & 3264 \\
\quad Batch Number Per GPU & 1 & 2 & 2 & 4 & 4 & 2 & 2 & 2 \\
\quad Max Episode Steps & 80 & 80 & 80 & 80 & 80 & 80 & 80 & 80 \\
\midrule

\rowcolor{lightgray}\multicolumn{9}{l}{\textit{\textbf{Training Hyperparameters}}} \\
\quad Rollout Epoch & 4 & 4 & 4  & 4  & 4 & 4 & 4 & 4 \\
\quad Update Epoch & 2 & 2 & 2  & 2  & 2 & 2 & 2 & 2 \\
\quad Micro Batch Size & 32 & 32 & 32 & 32 & 32 & 32 & 32 & 32 \\
\quad Global Batch Size & 2048 & 2048 &  2048 & 2048  & 2048 & 2048 & 2048 & 2048 \\
\midrule

\rowcolor{lightgray}\multicolumn{9}{l}{\textit{\textbf{Model Specifications}}} \\
\quad Model Num Step & 4 & 4 & 4 & 4 & 4 & 4 & 4 & 4 \\
\quad Chunk Size & 5 & 5 & 5 & 5 & 5 & 5 & 5 & 5 \\

\midrule
\rowcolor{lightgray}\multicolumn{9}{l}{\textit{\textbf{Dynamic Batching Settings}}} \\
\quad Max Batch Number & - & - & 1 & - & 1 & - & 1 & 1 \\
\quad Latency Threshold & - & - & -  &  - & - & - & - & - \\

\midrule
\rowcolor{lightgray}\multicolumn{9}{l}{\textit{\textbf{Asynchronous Training Settings}}} \\
\quad Rollout Async & Off & Off & On & Off & On & Off & On & On \\
\quad Train Async & Off & Off & Off & Off & Off & Off & Off & On \\

\midrule
\rowcolor{lightgray}\multicolumn{9}{l}{\textit{\textbf{Metric}}} \\
\quad Throughput & 191.51 & 187.76 & 170.61 & 178.76 & 155.82 & 143.10 & 154.40 & 255.58 \\

\bottomrule
\end{tabular*}
\end{table*}

\begin{table*}[!ht]
\centering
\caption{Detailed hyperparameters and throughput for different experimental setups on RoboCasa+$\pi_{0}$ with PPO.}
\label{tab:app_robocasa}
\footnotesize
\setlength{\tabcolsep}{2.2pt}
\renewcommand{\arraystretch}{1.2}
\resizebox{\linewidth}{!}{%
\begin{tabular}{@{}l*{9}{c}@{}}
\toprule
\textbf{Method} & Baseline & Baseline & \name\ & \name\ & Baseline & \name\ &  \name\ & \name\ & \name\ \\
\midrule

\rowcolor{lightgray}\multicolumn{10}{l}{\textit{\textbf{Placement Settings}}} \\
\quad Simulator & 0-7 & 0-7 & 0-7 & 0-7 & 0-5 & 0-5& 0-5 & 0-5 & 0-5 \\
\quad Generator & 0-7 & 0-7 & 0-7 & 0-7 & 6-7 & 6-7 & 6-7 & 6-7 & 6-7 \\
\quad Trainer & 0-7 & 0-7 & 0-7 & 0-7 & 0-7 & 0-7 & 0-7 & 0-7 & 0-7 \\
\midrule

\rowcolor{lightgray}\multicolumn{10}{l}{\textit{\textbf{Environment Settings}}} \\
\quad Total Env & 160 & 160 & 160 & 160 & 168 & 168 & 168 & 168 & 168  \\
\quad Batch Number Per GPU & 1 & 2 & 2 & 2 & 2 & 2 & 2 & 2 & 2 \\
\quad Max Episode Steps & 320 & 320 & 320 & 320 & 320 & 320 & 320 & 320 & 320 \\
\midrule

\rowcolor{lightgray}\multicolumn{10}{l}{\textit{\textbf{Training Hyperparameters}}} \\
\quad Rollout Epoch & 4 & 4 & 4  & 4  & 4 & 4 & 4 & 4 & 4 \\
\quad Update Epoch & 2 & 2 & 2  & 2  & 2 & 2 & 2 & 2 & 2 \\
\quad Micro Batch Size & 32 & 32 & 32 & 32 & 32 & 32 & 32 & 32 & 32 \\
\quad Global Batch Size & 1024 & 1024 &  1024 & 1024  & 1024 & 1024 & 1024 & 1024 & 1024\\
\midrule

\rowcolor{lightgray}\multicolumn{10}{l}{\textit{\textbf{Model Specifications}}} \\
\quad Model Num Step & 5 & 5 & 5 & 5 & 5 & 5 & 5 & 5 & 5 \\
\quad Chunk Size & 10 & 10 & 10 & 10 & 10 & 10 & 10 & 10 & 10 \\

\midrule
\rowcolor{lightgray}\multicolumn{10}{l}{\textit{\textbf{Dynamic Batching Settings}}} \\
\quad Max Batch Number & - & - & 1 & 1 & - & 1 & 2 & 1 & 2 \\
\quad Latency Threshold (ms) & - & - & -  &  - & - & - & 200 & - & 200 \\

\midrule
\rowcolor{lightgray}\multicolumn{10}{l}{\textit{\textbf{Asynchronous Training Settings}}} \\
\quad Rollout Async & Off & Off & On & On & Off & On & On & On & On \\
\quad Train Async & Off & Off & Off & On & Off & Off & On & Off & On \\

\midrule
\rowcolor{lightgray}\multicolumn{10}{l}{\textit{\textbf{Metric}}} \\
\quad Throughput & 409.6 & 379.25 & 437.60 & 505.68 & 332.92 & 355.64 & 349.32 & 454.63 & 455.01 \\

\bottomrule
\end{tabular}%
}
\end{table*}

\begin{table*}[!h]
\centering
\caption{Detailed hyperparameters and throughput for different experimental setups on LIBERO+GR00T N1.5 with PPO.}
\label{tab:app_libero}
\scriptsize

\renewcommand{\arraystretch}{1.25}

\begin{tabular*}{\textwidth}{@{\extracolsep{\fill}} l ccccccc @{}}
\toprule
\textbf{Method} & Baseline & Baseline & \name\ & \name\ & Baseline &  \name\ & \name\ \\
\midrule

\rowcolor{lightgray}\multicolumn{8}{l}{\textit{\textbf{Placement Settings}}} \\
\quad Simulator & 0-7 & 0-7 & 0-7 & 0-7 & 0-7 & 0-7 & 0-7 \\
\quad Generator & 0-7 & 0-7 & 0-7 & 0-7 & 0-7 & 0-7 & 0-7 \\
\quad Trainer & 0-7 & 0-7 & 0-7 & 0-7 & 0-7 & 0-7 & 0-7 \\
\midrule

\rowcolor{lightgray}\multicolumn{8}{l}{\textit{\textbf{Environment Settings}}} \\
\quad Total Env & 256 & 256 & 256 & 256 & 256 & 256 & 256 \\
\quad Batch Number Per GPU & 1 & 2 & 2 & 2 & 4 & 4 & 4  \\
\quad Max Episode Steps & 480 & 480 & 480 & 480 & 480 & 480 & 480 \\
\midrule

\rowcolor{lightgray}\multicolumn{8}{l}{\textit{\textbf{Training Hyperparameters}}} \\
\quad Rollout Epoch & 4 & 4 & 4  & 4  & 4 & 4 & 4  \\
\quad Update Epoch & 2 & 2 & 2  & 2  & 2 & 2 & 2 \\
\quad Micro Batch Size & 32 & 32 & 32 & 32 & 32 & 32 & 32 \\
\quad Global Batch Size & 1024 & 1024 & 1024 & 1024 & 1024 & 1024 & 1024 \\
\midrule

\rowcolor{lightgray}\multicolumn{8}{l}{\textit{\textbf{Model Specifications}}} \\
\quad Model Num Step & 4 & 4 & 4 & 4 & 4 & 4 & 4\\
\quad Chunk Size & 5 & 5 & 5 & 5 & 5 & 5 & 5 \\

\midrule
\rowcolor{lightgray}\multicolumn{8}{l}{\textit{\textbf{Dynamic Batching Settings}}} \\
\quad Max Batch Number & - & - & 1 & 1 & - & 1 & 1  \\
\quad Latency Threshold & - & - & -  & - & - & - & - \\

\midrule
\rowcolor{lightgray}\multicolumn{8}{l}{\textit{\textbf{Asynchronous Training Settings}}} \\
\quad Rollout Async & Off & Off & On & On & Off & On & On \\
\quad Train Async & Off & Off & Off & On & Off & Off & On \\

\midrule
\rowcolor{lightgray}\multicolumn{8}{l}{\textit{\textbf{Metric}}} \\
\quad Throughput & 217.27 & 242.30 & 297.38 & 349.01 & 240.13 & 344.47 & 423.39 \\

\bottomrule
\end{tabular*}
\end{table*}

\begin{table*}[!ht]
\centering
\caption{Detailed hyperparameters and throughput for different experimental setups on MetaWorld+$\pi_0$ with PPO.}
\label{tab:app_metaworld}
\footnotesize
\setlength{\tabcolsep}{2.2pt}
\renewcommand{\arraystretch}{1.2}
\resizebox{\linewidth}{!}{%
\begin{tabular}{@{}l*{10}{c}@{}}
\toprule
\textbf{Method} & Baseline & Baseline & \name\ &  \name\ & Baseline & \name\  &  \name\ & Baseline & \name\ & \name\ \\
\midrule

\rowcolor{lightgray}\multicolumn{11}{l}{\textit{\textbf{Placement Settings}}} \\
\quad Simulator & 0-7 & 0-7 & 0-7 & 0-7 & 0-7 & 0-7 & 0-7 & 0-5 & 0-5 & 0-5 \\
\quad Generator & 0-7 & 0-7 & 0-7 & 0-7 & 0-7 & 0-7 & 0-7 & 6-7 & 6-7 & 6-7 \\
\quad Trainer & 0-7 & 0-7 & 0-7 & 0-7 & 0-7 & 0-7 & 0-7 & 0-7 & 0-7 & 0-7 \\
\midrule

\rowcolor{lightgray}\multicolumn{11}{l}{\textit{\textbf{Environment Settings}}} \\
\quad Total Env & 512 & 512 & 512 & 512 & 512 & 512 & 512 & 768 & 768 & 768 \\

\quad Batch Number Per GPU & 1 & 2 & 2 & 2 & 4 & 4 & 4 & 2 & 2 & 2 \\
\quad Max Episode Steps & 100 & 100 & 100 & 100 & 100 & 100 & 100 & 100 & 100 & 100 \\
\midrule

\rowcolor{lightgray}\multicolumn{11}{l}{\textit{\textbf{Training Hyperparameters}}} \\
\quad Rollout Epoch & 4 & 4 & 4  & 4  & 4 & 4 & 4 & 4 & 4 & 4\\
\quad Update Epoch & 2 & 2 & 2  & 2  & 2 & 2 & 2 & 2 & 2 & 2\\
\quad Micro Batch Size & 128 & 128 & 128 & 128 & 128 & 128 & 128 & 128 & 128 & 128\\
\quad Global Batch Size & 2048 & 2048 &  2048 & 2048  & 2048 & 2048 & 2048 & 2048 & 2048 & 2048 \\
\midrule

\rowcolor{lightgray}\multicolumn{11}{l}{\textit{\textbf{Model Specifications}}} \\
\quad Model Num Step & 4 & 4 & 4 & 4 & 4 & 4 & 4 & 4 & 4 & 4\\
\quad Chunk Size & 5 & 5 & 5 & 5 & 5 & 5 & 5 & 5 & 5 & 5\\

\midrule
\rowcolor{lightgray}\multicolumn{11}{l}{\textit{\textbf{Dynamic Batching Settings}}} \\
\quad Max Batch Number & - & - & 1 & 1 & - & 1 & 1 & - & 1 & 1 \\
\quad Latency Threshold & - & - & -  & - & - & - & - & - & - & - \\

\midrule
\rowcolor{lightgray}\multicolumn{11}{l}{\textit{\textbf{Asynchronous Training Settings}}} \\
\quad Rollout Async & Off & Off & On & On & Off & On & On & Off & On & On \\
\quad Train Async & Off & Off & Off & On & Off & Off & On & Off & Off & On \\

\midrule
\rowcolor{lightgray}\multicolumn{11}{l}{\textit{\textbf{Metric}}} \\
\quad Throughput &130.01 &143.97 &165.54 &209.81 &144.70 &197.27 &268.00 &92.29 &113.62 &166.94 \\
\bottomrule
\end{tabular}%
}
\end{table*}

\begin{figure}[t]
    \centering
    \includegraphics[width=\linewidth]{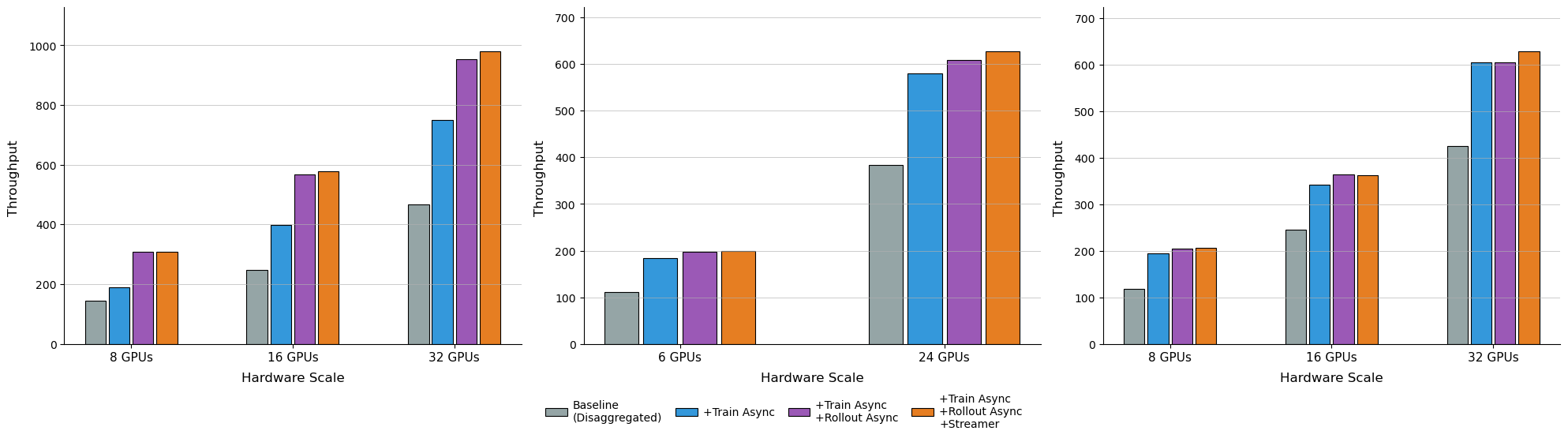}
    \caption{LIBERO + OpenVLA-OFT. The left, center, and right panels represent rollouts with GPU allocation ratios of 1:1, 2:1, and 3:1 between rollout workers (simulator and generator) and trainer, respectively.}
    \label{fig:combine_dis_b}
\end{figure}

\subsection{Implementation Details and Additional Throughput Results}\label{app:experiments-implementation}

Tables~\ref{tab:app_maniskill}--\ref{tab:app_metaworld} report full per-run settings and measured throughput for the four simulators used in our study, complementing the summary in Figure~\ref{fig:throughput_comparison}. 
The bold values in the tables are the ones we reported in Figure~\ref{fig:throughput_comparison}. 
All experiments in this section were conducted on a single node with 8 GPUs.
We sweep Colocated versus Hybrid placement, the number of environment batches hosted per Simulator GPU, and which asynchronous features are enabled (rollout-side decoupling with dynamic batching, training-side streaming optimization, or both).
Each value is an average over $5$ training steps.
We present a detailed analysis below.

\paragraph{ManiSkill ($\pi_{0.5}$, GRPO).}
ManiSkill~\citep{mu2021maniskill} combines very large vectorized environment counts with GPU-accelerated simulation.
As shown in Table~\ref{tab:app_maniskill}, we can set a very large environment batch size: up to $256$ ($2048/8$) for colocated placement and $272$ ($3276/(6*2)$) for hybrid placement. 
Under colocated placement, it is interesting to see that the throughput of the synchronous baseline is already the best. 
The reason is that the generator's throughput already reaches its peak with respect to batch size. 
Therefore, there is no benefit from increasing the batch number and aggregating different environment batches for one generator inference. 
In this case, the increased context-switching overhead from adding asynchrony to the rollout phase dominates any benefit from the asynchronous rollout design. 
On the other hand, in hybrid placement, both rollout asynchrony and train asynchrony from \name\ improve throughput. Notably, the highest throughput is achieved when both are enabled. 
The improvement is about 78.6\% over the synchronous hybrid baseline and 33.5\% over the synchronous colocated baseline.

\paragraph{RoboCasa ($\pi_0$, PPO).}
RoboCasa~\citep{nasiriany2024robocasa} is a simulation environment focused on manipulation tasks in diverse visual scenes, producing high-resolution observations. 
Although the GPU memory occupation of a single RoboCasa environment is low, the computational cost of physics simulation is resource-intensive, making it difficult to scale up the environment batch size per GPU. 
As reported in Table~\ref{tab:app_robocasa}, we configured a total of 160 environments for colocated placement and 168 for hybrid placement.
In the colocated setup, the best synchronous baseline achieves 409.6 with a single environment batch per GPU. 
Simply increasing the number of batches decreases throughput to 379.25. 
However, \name\ improves throughput by 15.4\% with rollout asynchrony and 33.3\% with full asynchrony.
For hybrid placement, the benefit of enabling training asynchrony is more significant. 
However, the throughput of \name\ with hybrid placement is lower than with colocated placement. 
This is again due to the characteristics of RoboCasa: GPU memory occupation is relatively low, but the number of environments is limited by CPU resources. 
Thus, having each GPU host all three resource groups makes better use of GPU memory and reduces communication overhead between GPUs.

\paragraph{LIBERO (GR00T N1.5, PPO).}
LIBERO~\citep{liu2023libero} is representative of CPU-heavy MuJoCo-style stepping.
Therefore, all runs in Table~\ref{tab:app_libero} use colocated placement on GPUs, isolating how batching and asynchrony interact when the simulator, generator, and trainer time-share the same devices.
We mainly vary the number of environment batches per GPU while maintaining the same total environment count of 256.
As shown in the table, since the baseline training adopts a synchronous pipeline, increasing the batch number per GPU does not yield a clear improvement in throughput (217.27 for 1, 242.30 for 2, and 240.13 for 4).
However, by introducing the asynchronous rollout design, \name\ significantly increases throughput by 22.7\% and 43.5\% with 2 and 4 batches per GPU, respectively.
This means that inference time can be effectively hidden by simulator stepping time, consistent with the fact that stepping time is much longer than inference time in MuJoCo-style environments. 
When we further activate training asynchrony, throughput can be improved to 423.39, which is about a 74.7\% improvement over the best synchronous performance.

\paragraph{Meta-World ($\pi_0$, PPO).}
Meta-World~\citep{yu2020meta} is more parallelizable than LIBERO and RoboCasa. 
Therefore, we can set a larger total environment count (512 for colocated and 768 for hybrid). 
As shown in Table~\ref{tab:app_metaworld}, Meta-World exhibits similar trends to LIBERO in colocated placement: the synchronous baseline does not benefit from splitting environments into multiple batches, but \name\ significantly improves throughput by activating rollout and training asynchrony.
The best configuration uses 4 batches per GPU with full asynchrony, achieving 268 in throughput, about an 85.2\% improvement over the best synchronous performance. 
Moreover, for hybrid placement, the throughput of \name\ achieves about an 80.9\% improvement over the synchronous baseline, showing that asynchronous design also helps mitigate communication overhead and rebalancing costs in hybrid placement.

\subsection{Ablation on Different Ratios of GPU Allocation}\label{app:experiments-placement}

As shown in the detailed results in Section~\ref{app:experiments-implementation}, the GPU resource ratio for hybrid placement is 3:1 between the simulator and the generator, i.e., for a total of 8 GPUs, we let 6 GPUs load environments and 2 GPUs load a model. 
Apart from this 3:1 ratio, we also evaluate the performance of \name\ under 1:1 and 2:1 ratios. 
We use a different hybrid placement here for the LIBERO environment: we separate all GPUs into rollout workers and the trainer. 
For example, with a 1:1 ratio on an 8-GPU node, we let 4 GPUs simultaneously load an environment and a model, while the other 4 GPUs load only a model for training. 
Figure~\ref{fig:combine_dis_b} presents the results for this placement on LIBERO + OpenVLA-OFT with different ratios; the left, center, and right panels correspond to rollout-side allocation ratios of 1:1, 2:1, and 3:1 between rollout workers and the trainer. 
Furthermore, we decouple full asynchrony into train async and rollout async, and evaluate how different levels of asynchrony influence throughput.
Train async is further decoupled into basic training asynchrony and streamer training: basic training still waits for a full epoch of rollout data before starting training, while streamer training can start immediately after receiving the first batch of rollout data.

As shown in the figure, different placement ratios show similar trends, which is also consistent with our analysis in Section~\ref{sec:experiments-ablation}: train async tends to yield a larger step-change in throughput than rollout async, and the combination of both yields the best performance. 
The improvement from rollout async is more significant at the 1:1 ratio than at others. 
This is because the 1:1 ratio allocates fewer GPU resources to rollout workers, whereas asynchrony helps better utilize GPU resources by overlapping inference time with stepping time. 
Streamer training alone does not show significant improvement because the current rollout trajectory has a fixed length, so all training data from the same rollout epoch arrives at roughly the same time.

\clearpage
\subsection{Additional Success Rate Curve}

\begin{figure}[t]
    \centering
    \includegraphics[width=0.32\linewidth]{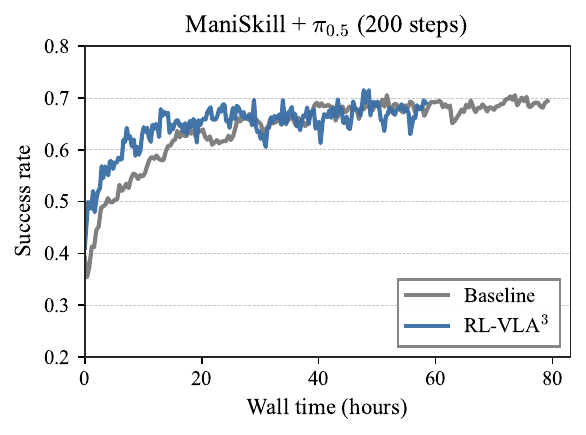}\hfill
    \includegraphics[width=0.32\linewidth]{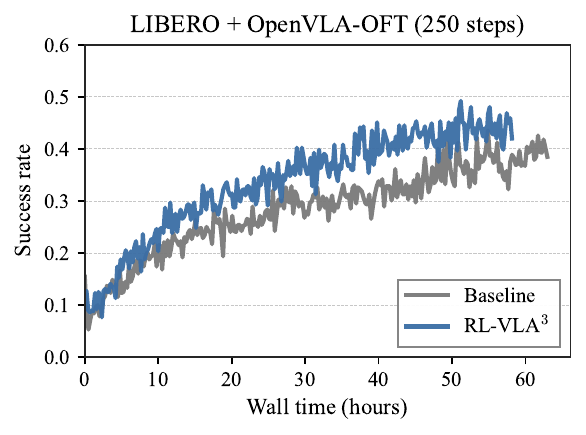}
    \includegraphics[width=0.32\linewidth]{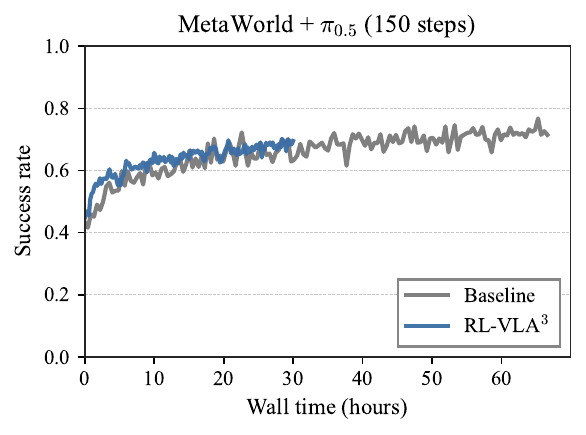}
    \caption{Success-rate curves for ManiSkill+$\pi_{0.5}$, LIBERO+OpenVLA-OFT, and Meta-World+$\pi_{0.5}$ under the synchronous baseline versus \name. The trajectories are aligned by environment steps, confirming comparable sample efficiency, while \name's higher throughput substantially reduces the wall-clock time needed to reach comparable success levels.}
    \label{fig:success_rate_wall_clock}
\end{figure}

We present an alternative version of Figure~\ref{fig:success_rate}, where we change the x-axis from training steps to wall-clock time. 
As shown in Figure~\ref{fig:success_rate_wall_clock}, under the same configuration, \name\ significantly improves training speed without noticeably compromising training stability or convergence compared to the baseline.

\subsection{More Metrics: CPU/GPU Utilization}\label{app:resource-utilization}

In this section, we present additional evaluation metrics, including CPU utilization, RAM usage, GPU utilization, and GPU memory usage. 
We plot the entire \emph{first} training step, so the elapsed times differ. 
From the usage curves, it becomes clearer how \name's implementation improves throughput over the synchronous baseline.
A representative LIBERO case is discussed in Section~\ref{sec:experiments-resource}; here we provide the remaining environments and extended discussion.

\paragraph{ManiSkill ($\pi_{0.5}$, GRPO).}

\begin{figure}
    \centering
    \includegraphics[width=0.9\linewidth]{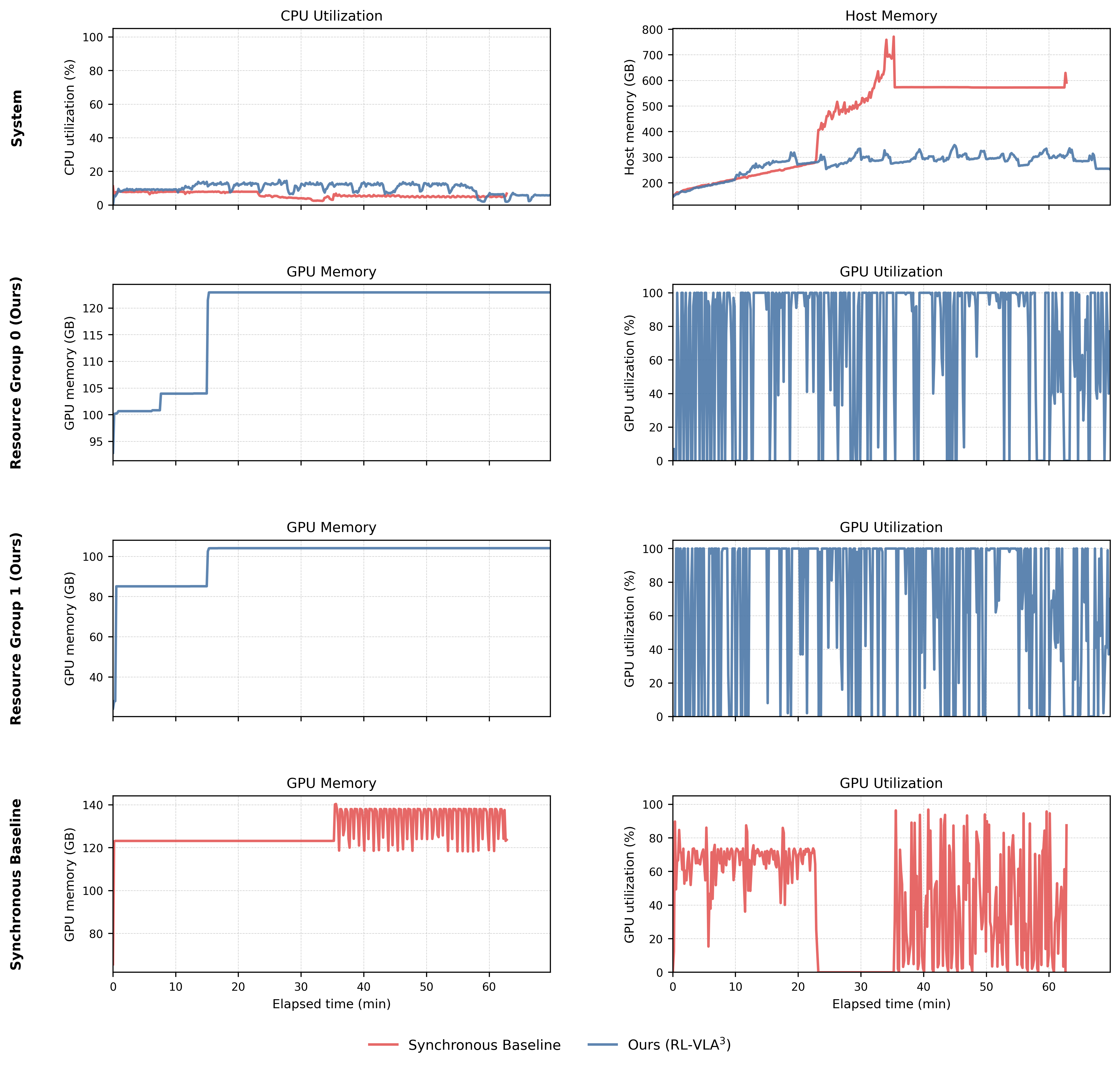}
    \caption{Comparison of resource utilization during the first step between the Colocated (Baseline) and Hybrid (Ours) on ManiSkill+$\pi_{0.5}$. Note that in this step, Hybrid (Ours) processes approximately 1.5$\times$ the data volume of Colocated (Baseline), resulting in a longer elapsed time.}
    \label{fig:resource-maniskill}
\end{figure}

For ManiSkill, we select two representative cases: colocated placement for the baseline (first column of Table~\ref{tab:app_maniskill}) and hybrid placement for \name\ (last column of Table~\ref{tab:app_maniskill}), which achieve the highest throughput under each method. 
For the Hybrid placement, the total resources are divided into two resource groups: each GPU in resource group 0 hosts one Simulator and one Trainer, while each GPU in resource group 1 hosts one Generator and one Trainer. 
For the colocated placement, each GPU hosts one Simulator, one Generator, and one Trainer.

For \name, the noticeable increase in GPU memory usage arises because the data from the first rollout enters the training engine. Due to asynchronous training and inference, the GPU has already started training, thereby increasing memory consumption. 
In contrast, for the baseline, training starts only after all rollout data have been collected, so the peak GPU memory usage occurs much later.
Moreover, we observe that because the baseline transfers all data in one batch after collection is complete, there is a pronounced spike in host memory usage and GPU waiting at around 23 minutes. This happens because the training data of the VLA includes large inputs such as images, which consume considerable memory and time when the dataset is large.
Asynchronous training–inference means we transfer data in batches; after each training step the data can be discarded (following pre-defined rules), which reduces the burden on the system.

\paragraph{LIBERO (GR00T N1.5, PPO) and Meta-World ($\pi_0$, PPO).}

We discuss LIBERO and Meta-World together because both selected cases use colocated placement, allowing us to compare resource utilization between the synchronous baseline and \name. The specific configurations are those that achieve the highest throughput for the Baseline and \name\ in Table~\ref{tab:app_libero} and Table~\ref{tab:app_metaworld}, respectively.

\begin{figure}
    \centering
    \includegraphics[width=0.9\linewidth]{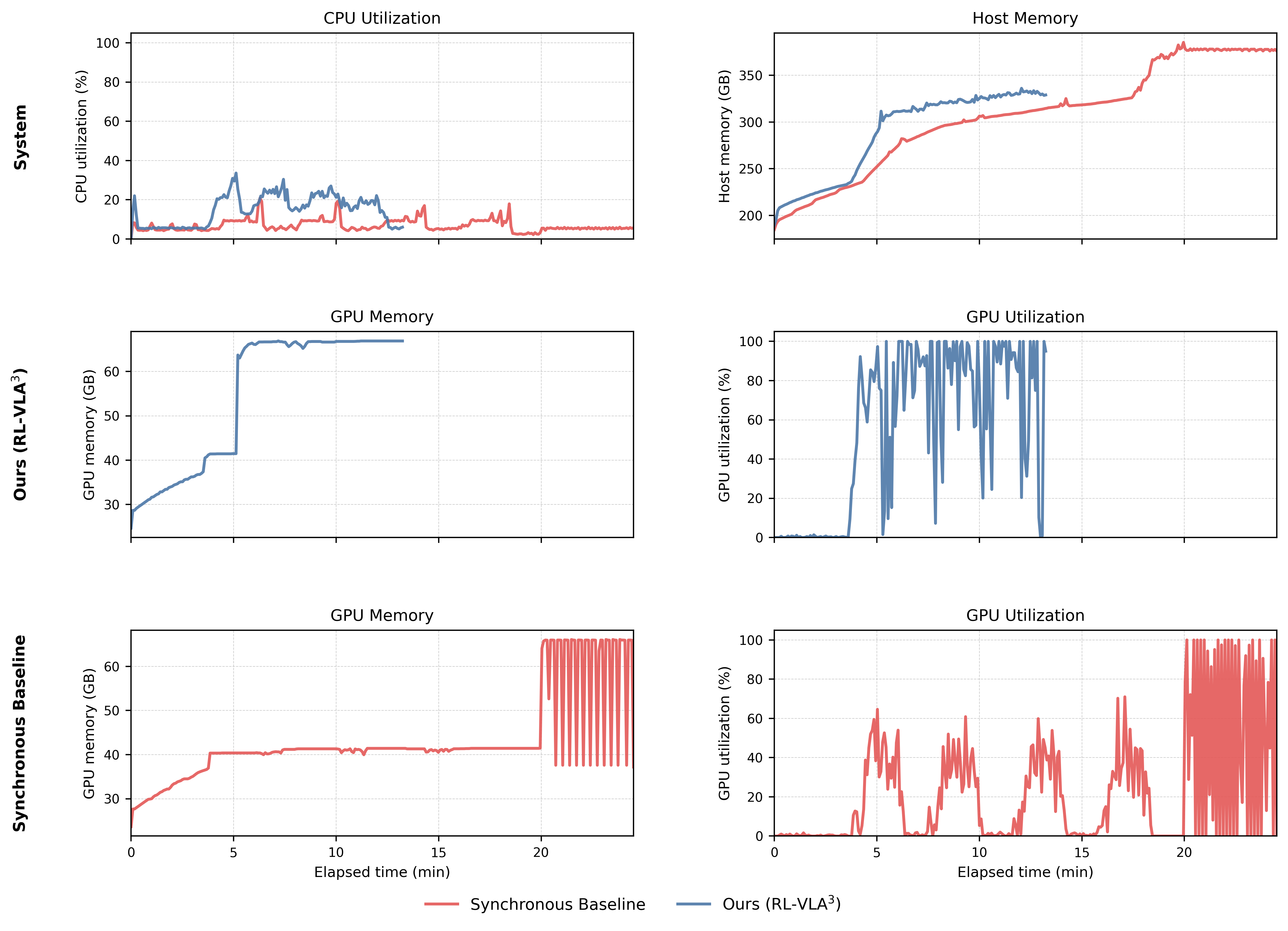}
    \caption{Comparison of resource utilization during the first step between Colocated (Baseline) and Colocated (Ours) on MetaWorld+$\pi_{0.5}$.}
    \label{fig:resource-metaworld}
\end{figure}

As can be seen, Figures~\ref{fig:resource-libero} and~\ref{fig:resource-metaworld} exhibit very similar comparative effects. We can clearly observe that \name\ starts training immediately after the first rollout finishes, rather than waiting for all rollouts to complete.
At the same time, similar to the ManiSkill case, asynchronous training–inference with batched data transfer reduces the high GPU waiting time and high CPU load caused by one-shot transfers.
Another interesting observation is that, for the Meta-World baseline strategy, we see noticeable GPU idle periods between the four rollout epochs. This is likely because we increased the number of parallel environments per rollout to boost throughput, making the initialization phase very time-consuming. This suggests that we could leverage the environment initialization phase to use GPU resources more efficiently for training, thereby improving throughput.

\end{document}